\documentclass[10pt,twocolumn,letterpaper]{article}

\usepackage{cvpr}
\usepackage{times}  
\usepackage{helvet}  
\usepackage{courier}  
\usepackage{epsfig}
\usepackage{amsmath}
\usepackage{amssymb}
\usepackage{multirow}
\usepackage{graphicx}  
\usepackage{booktabs}
\usepackage{epsfig,epstopdf,graphicx,subfigure}
\usepackage{amsmath,amssymb,amscd,epsfig,amsfonts,rotating}
\usepackage{booktabs} 
\usepackage{xcolor}
\usepackage{algpseudocode}
\usepackage{algorithm,tabularx}
\makeatletter
\newcommand{\multiline}[1]{%
  \begin{tabularx}{\dimexpr\linewidth-\ALG@thistlm}[t]{@{}X@{}}
    #1
  \end{tabularx}
}
\makeatother
\newcommand{\bs}{\boldsymbol}
\newcolumntype{C}[1]{>{\centering\arraybackslash}p{#1}}
\cvprfinalcopy 

\def\cvprPaperID{2847} 

\begin{document}

\title{Point Cloud Completion by Skip-attention Network with Hierarchical Folding}

\author{Xin Wen\textsuperscript{1}, Tianyang Li\textsuperscript{1}, Zhizhong Han\textsuperscript{2}, Yu-Shen Liu\textsuperscript{1}\thanks{Corresponding author. This work was supported by National Key R\&D Program of China (2018YFB0505400).}\\
\textsuperscript{1}School of Software, BNRist, Tsinghua University, Beijing, China\\
\textsuperscript{2}Department of Computer Science, University of Maryland, College Park, USA\\
{\tt\small\{x-wen16,lity16\}@mails.tsinghua.edu.cn\hspace{1mm}
h312h@umd.edu\hspace{1mm}
liuyushen@tsinghua.edu.cn}
}

\maketitle
\thispagestyle{empty}

\begin{abstract}
    Point cloud completion aims to infer the complete geometries for missing regions of 3D objects from incomplete ones.
    Previous methods usually predict the complete point cloud based on the global shape representation extracted from the incomplete input. However, the global representation often suffers from the information loss of structure details on local regions of incomplete point cloud.
    To address this problem, we propose Skip-Attention Network (SA-Net) for 3D point cloud completion. Our main contributions lie in the following two-folds.
    First, we propose a skip-attention mechanism to effectively exploit the local structure details of incomplete point clouds during the inference of missing parts.
    The skip-attention mechanism selectively conveys geometric information from the local regions of incomplete point clouds for the generation of complete ones at different resolutions, where the skip-attention reveals the completion process in an interpretable way.
    Second, in order to fully utilize the
    selected geometric information encoded by skip-attention mechanism at different resolutions, we propose a novel structure-preserving decoder with hierarchical folding for complete shape generation.
    The hierarchical folding preserves the structure of complete point cloud generated in upper layer by progressively detailing the local regions, using the skip-attentioned geometry at the same resolution.
    We conduct comprehensive experiments on ShapeNet and KITTI datasets, which demonstrate that the proposed SA-Net outperforms the state-of-the-art point cloud completion methods.
\end{abstract}

\section{Introduction}
Recently, point cloud has received an extensive attention as a format of 3D objects, which can be easily accessed by 3D scanning devices and depth cameras. However, the raw point clouds produced by those devices are usually sparse, noisy and mostly with serious missing regions due to the limited view angles or occlusion \cite{yuan2018pcn}, which are difficult to be directly processed by the further shape analysis/rendering methods.
Therefore, raw point cloud  preprocessing becomes an important requirement for many real-world 3D computer vision applications.
In this paper, we focus on the task of completing the missing regions of 3D shapes represented by point clouds.


\begin{figure}[!t]
  \centering
  \includegraphics[width=\columnwidth]{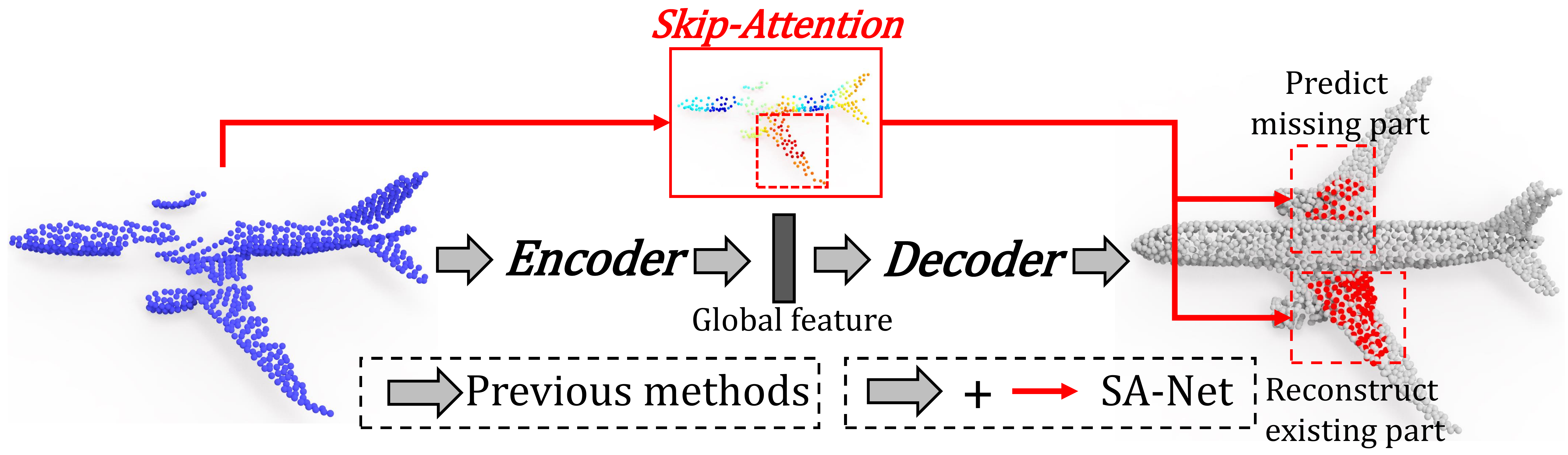}
  \caption{Illustration of the proposed skip-attention. Compared with the previous methods that simply rely on the global shape representation for completing point clouds, our skip-attention mechanism directly searches for informative local regions in input airplane shape, and selectively uses these regions for predicting the missing right wing or reconstructing the similar left wing (red).
  }
  \label{fig:plane_intro}
\end{figure}

The task of point cloud completion can be roughly decomposed into two targets \cite{tchapmi2019topnet,yuan2018pcn}. The first target is to preserve the geometric shape information of the original input point cloud, and the second target is to recover the missing regions according to the given inputs.
In order to achieve such two targets, current studies usually followed the paradigm framework to learn a global shape representation from incomplete point clouds, which is further leveraged to estimate the missing geometric information \cite{yang2018foldingnet,yuan2018pcn,hu20193d}.
However, the encoded global shape representation often suffers from the information loss of some structure details on local regions of incomplete point clouds, which should be fully preserved for further inferring the missing geometric information.
As shown in Figure \ref{fig:plane_intro}, to predict the complete wings of an airplane, the network should first preserve the existing left wing in the incomplete point cloud. And then, in order to infer the missing right wing, the network could refer to the existing left wing according to the pattern similarity between the regions of two similar wings.

An intuitive idea to address this problem is to adopt the \emph{skip-connection} mechanism like U-Net \cite{ronneberger2015u}, which is widely used for local region reconstruction and reasoning in images. However, there are two problems for directly adopting skip-connection into point cloud completion. First, the previous skip-connection developed in \cite{ronneberger2015u} can not be directly applied to unordered inputs, since it concatenates the feature vectors according to the pixel order of 2D grids.
Second, in the task of point cloud completion, not all the local region features under each level of resolutions will be of help for shape inferring and reconstruction. Equally revisiting them with skip-connection may introduce the information redundancy, and limit the feature learning ability of the entire network.

Therefore, in order to preserve the information of structure details while addressing the problem of skip-connection, we propose a novel deep neural network for point cloud completion, named \emph{Skip-Attention Network} (SA-Net). The network is designed in an end-to-end framework, where an encoder-decoder architecture is specially designed for feature extraction and shape completion.
The skip-attention refers to the attention based feature pipeline, which reveals completion process in an interpretable way.
The skip-attention selectively conveys geometric information from the local regions of incomplete point clouds for the generation of the complete ones at different resolutions.
The skip-attention enables the decoder to fully exploit and preserve the structure details on local regions. Compared with the skip-connection, the skip-attention can be generalized to unordered point clouds, since attention mechanism has no pre-requirements on the order of the input features. Moreover, our skip-attention provides an attentional choice for network to revisit the features under different resolutions, which allows the network to selectively incorporate the features encoded with desirable geometric information, and avoid the problem of information redundancy.


In order to fully utilize the selected geometric information from skip-attention at different resolutions, we further propose a structure-preserving decoder with \emph{hierarchical folding} to generate complete point clouds. The hierarchical folding preserves the structure of point cloud generated in the upper layer, by progressively detailing the local regions using the skip-attentioned geometric information at the same resolution from the encoder.
Specifically, the decoder has the same number of resolution levels as the encoder, with the skip-attention connecting each level of encoder to the corresponding level of decoder. In order to hierarchically fold the point clouds through levels, we propose to sample 2D grids with an increasing density from a 2D plane of fixed size.
Compared with the decoders in existing point cloud completion methods \cite{yuan2018pcn,tchapmi2019topnet,yang2018foldingnet}, the proposed structure-preserving decoder can preserve the structure details on local regions under the whole resolution levels, which enables the network to predict complete shape that maintains the global shape consistency while capturing more local region information.
Our main contributions can be summarized as follows.
\begin{itemize}
  \item We propose a novel Skip-Attention Network (SA-Net) for the point cloud completion task, which achieves state-of-the-art results.
  Moreover, the architecture of SA-Net can also be used for improving the performance of shape segmentation, and achieving the state-of-the-art results in unsupervised shape classification.
  \item We propose the skip-attention mechanism to fuse the informative local region features from encoder into the point features of decoder at different resolutions, which enables the network to infer the missing regions using more detailed geometry information from incomplete point clouds. In addition, skip-attention reveals the completion process in an interpretable way.
  \item We propose a structure-preserving decoder for high quality point cloud generation. It can progressively detail the point clouds at different resolutions with hierarchical folding, which hierarchically preserves the structure of complete shape at different resolutions.
\end{itemize}


\begin{figure*}[!t]%
  \centering
  \includegraphics[width=\textwidth]{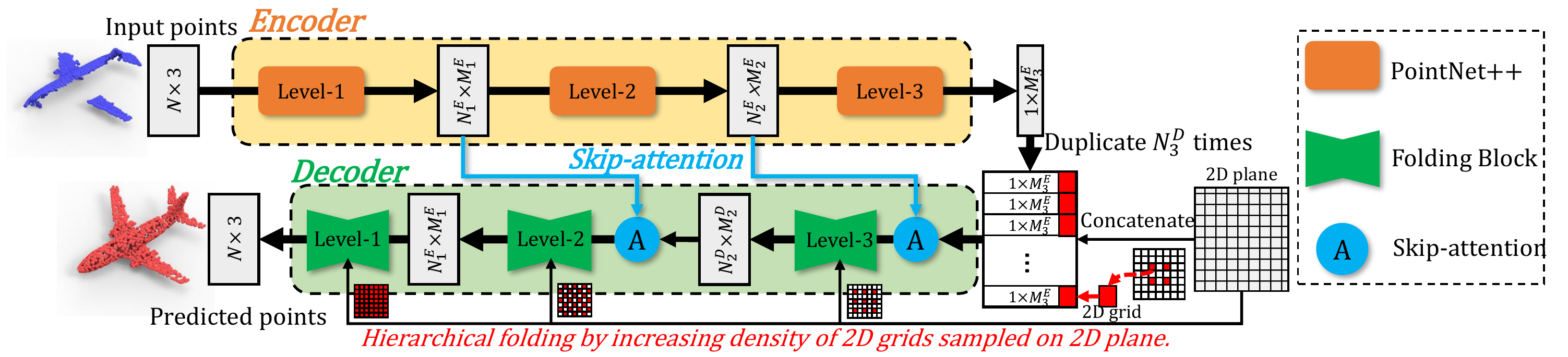}
  \caption{The overall architecture of SA-Net. SA-Net mainly consists of three modules: the encoder (yellow) aims to extract local region features from the input point clouds; the structure-preserving decoder (green) aims to reconstruct the complete point clouds and preserve the local region details; the skip-attention (sky blue) bridges the local region features in encoder and the point features in decoder.
  }
  \label{fig:overview}
\end{figure*}

\section{Related Work}
3D computer vision is an active research field in recent year \cite{gao2015query,HanCyber17a,Zhizhong2016,Zhizhong2016b,liu2019sequence,han2017boscc,liu2011computing}, where the studies of 3D shape completion lead to many branches. For examples, geometry based \cite{sung2015data,berger2014state,thanh2016field,wei2019local} methods exploit the geometric features of surface on the partial input to generate the missing part of 3D shapes, and alignment-based methods \cite{shao2012interactive,kalogerakis2012probabilistic,martinovic2013bayesian,shen2012structure} maintain a shape database and search for the similar patches to fill the incomplete regions of 3D shapes. Our method belongs to the deep learning based methods, which benefits a lot from the recent development of deep neural network in 3D computer vision \cite{han2019parts4feature,liu2020lrc,han20193dviewgraph,han2019view,han20182seq2seq,han20193d2seqviews,han2018seqviews2seqlabels,han2018deep}. This branch can be further categorized according to the input form of 3D shapes.

\textbf{Volumetric shape completion.}
3D volumetric shape completion is a direction that is benefited a lot from the progress in 2D computer vision.
Notable work like 3D-EPN \cite{dai2017shape} considered a progressive reconstruction of 3D volumetric shapes.
And Han et. al \cite{han2017high} combined the inference of global structure with the local geometry refinement to directly generate the complete 3D volumetric shape of high resolution. More recently, the variational auto-encoder was introduced to learn a shape prior for inferring the latent representation of complete shapes \cite{stutz2018learning}. Although fascinating improvements have been made in the research area of 3D volumetric data, the computational cost which is cubic to the resolution of input data makes it difficult to process fine-grained shapes.

\textbf{Point cloud completion.}
The point cloud based 3D shape completion is a surging research area benefited from the pioneering work of PointNet \cite{qi2017pointnet} and PointNet++ \cite{qi2017pointnet++}. As a compact representation of 3D shapes, point cloud can represent arbitrary detailed structure of 3D shape with a smaller storage cost compared to 3D volumetric data. Recent notable studies like PCN \cite{yuan2018pcn}, FoldingNet \cite{yang2018foldingnet} and AtlasNet \cite{groueix2018atlasnet} usually learn a global representation from partial point cloud, and generate the complete shape based on the learned global feature. Following the same practice, a tree-structured decoder was proposed in TopNet \cite{tchapmi2019topnet} for better structure-aware point cloud generation. By combining the reinforcement learning with the adversarial network, RL-GAN-Net \cite{sarmad2019rl} and Render4Completion \cite{hu2019render4completion} further improved the reality and consistency of the generated complete point cloud with the ground truth. However, most of these studies suffer from the information loss of structure details, as they predict the whole point cloud only from a single global shape representation.


\section{The Architecture of SA-Net}
Figure \ref{fig:overview} shows the overall architecture of SA-Net, which consists of an encoder and a structure-preserving decoder. Between the encoder and the decoder, the skip-attention serves as the pipeline that connects the local region features (extracted from different resolutions in encoder) with the point features in the corresponding resolutions of decoder.

\subsection{Encoder}
Given the input point cloud of size $N$=2,048 with its 3-dimensional coordinates, the encoder of SA-Net aims to extract the features from the incomplete input point clouds. In SA-Net, we adopt the PointNet++ \cite{qi2017pointnet++} framework as the backbone of our point cloud feature encoder. As shown in Figure \ref{fig:overview}, there are three levels of feature extraction, with the first level and the second level sampling the input point cloud into the size $N^E_1$=512 and $N^E_2$=256 (the superscript $E$ denotes the encoder), and the last level grouping the input point cloud into a global representation. As a result, the encoder generates one global representation, and some local region features extracted from different resolution levels for the input point clouds, respectively.

\subsection{Structure-Preserving Decoder}
Considering that the encoder extracts the local region features from different resolution levels, it is a natural practice for decoder to generate point features following the same way but with inverse resolution levels. This allows the skip-attention to establish a level-to-level connection between the extracted local region features in encoder and the generated point features in decoder. Inspired by this idea, we propose the structure-preserving decoder, which aims to progressively generate the complete point clouds and preserve the structure details of local regions under all resolution levels.
\begin{figure}[!t]
  \centering
  \includegraphics[width=.9\columnwidth]{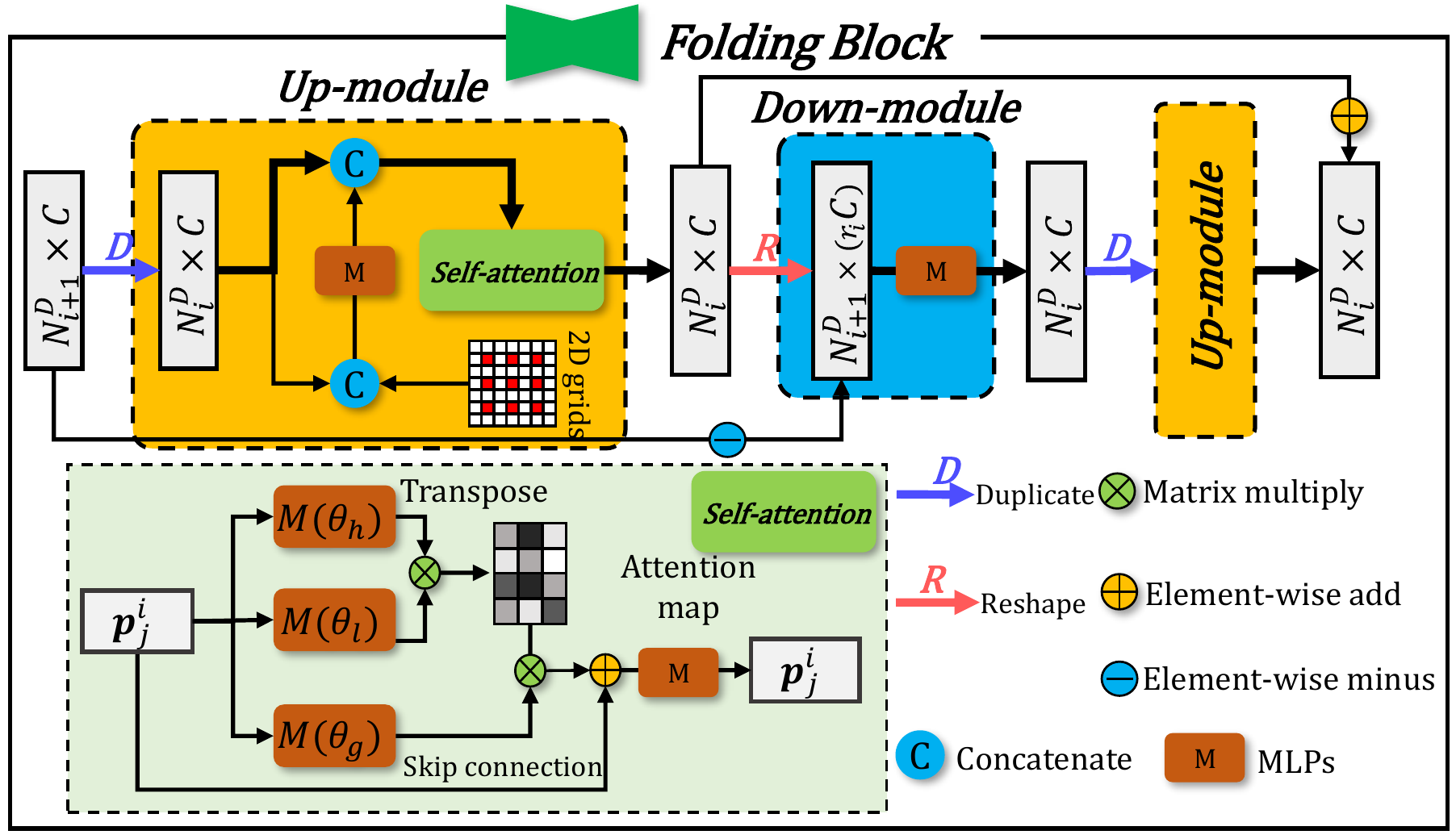}
  \caption{Illustration of the folding block, which consists of a down-module and two up-modules with self-attention inside. Folding block aims to lift the number of point features and refine the geometric information lying within these features.
  }
  \label{fig:expansion_module}
\end{figure}
Specifically, as shown in Figure \ref{fig:overview}, the structure-preserving decoder hierarchically folds the point clouds for three resolution levels, which is equal to the number of resolution levels in the encoder. Each resolution level of decoder consists of a skip-attention to convey the local region features from the same level of encoder, and a folding block to increase the number of point features.

\subsection{Folding Block}
Except for lifting the number of point features, the folding block also concerns the refinement of the expanded point features, which allows the decoder to produce more consistent geometric details on the local regions of point clouds.
Note that, such problem is usually ignored by previous methods, in which they either directly fold the entire point set based on the duplicated global representation \cite{yuan2018pcn,yang2018foldingnet}, or simply produce the point clouds through multi-layer perceptrons (MLPs) and reshape operations \cite{tchapmi2019topnet}.
In SA-Net, we take the inspiration of the up-down-up framework from \cite{li2019pu} to address this problem, which is adopted as the base of our folding block.
Figure \ref{fig:expansion_module} shows the detailed structure of the folding block in the $i$-th level of decoder.

\textbf{The up-module with hierarchical folding.} As shown in the yellow part of Figure \ref{fig:expansion_module}, for the input $N^D_{i+1}$ (the superscript $D$ denotes the decoder) point features from the previous level, the up-module first copies the point features by the time of up-sampling ratio $r_i=\frac{N^D_{i}}{N^D_{i+1}}$, and concatenates them with the 2D grids. Different from previous folding based decoders \cite{li2019pu,yuan2018pcn,yang2018foldingnet}, which only have one resolution level for point cloud generation, the decoder in SA-Net progressively generates the point clouds for multiple resolution levels. In order to hierarchically fold point clouds through these levels, we propose to sample 2D grids with an increasing density from the 2D plane of fixed size.
Specifically, for the $N_i^D$ point features in the $i$-th level of decoder, the $N_i^D$ 2D grids is evenly sampled from the $46\times 46$ 2D plane (the smallest number of square greater than 2,048), as illustrated in the up-module of Figure \ref{fig:expansion_module}.
These sampled 2D grids are then concatenated with the point features. After that, the point features with 2D grids are passed through MLPs and transformed into 3-dimensional latent codewords \cite{yang2018foldingnet}. These 3-dimensional codewords are again concatenated with the point features in the $i$-th level of decoder.

In order to integrate the semantic and spatial relationships between these point features, we adopt a self-attention module with MLPs to establish the inner links between features, which aims to selectively fuse the similar features together through attention mechanism. This process is shown in the bottom half of Figure \ref{fig:expansion_module}.
Given the $j$-th point feature $\bs{p}^i_j$ of $i$-th level in decoder, the skip-connection first calculates the attention scores $\{a_{j,k}| k=1,2,...,N^D_i\}$ between $\bs{p}^i_j$ and all of the point features $\{\bs{p}^i_k | k=1,2,...,N^D_i\}$ in $i$-th level of decoder as
\begin{equation}\small
  a_{j,k} = \frac{\exp({\rm M}(\bs{p}^i_j|\theta_h)^{\rm T}\cdot {\rm M}(\bs{p}^i_k|\theta_l))}{\sum_{n=1}^{N_i}\exp({\rm M}(\bs{p}^i_j|\theta_h)^{\rm T}\cdot {\rm M}(\bs{p}^i_n|\theta_l))}
\end{equation}
where ${\rm M}$ denotes the MLPs with parameters $\theta$, and $\rm T$ denotes the transposition operation. $h$ and $l$ indicate that two MLPs have different parameters. We take the weighted sum of point features $\{\bs{p}^i_k\}$ as the final context vector, and fuse it into the point feature $\bs{p}^i_j$ as follow:
\begin{equation}\small
\label{eq:fuse}
  \bs{p}^i_j \leftarrow  \bs{p}^i_j + \sum_{k=1}^{N^D_i}a_{j,k}\cdot {\rm M}(\bs{p}^i_k|\theta_g)
\end{equation}

\textbf{The down-module.} The point features expanded by the up-module actually occupy a small local region in the feature space, which can be aggregated as one local region feature through reshape and feature concatenation. Such aggregated local region feature can be regarded as a refined point feature of higher quality compared with the one in previous levels, since it contains not only the information from previous level of decoder, but also the detailed information produced by the current up-module. Then, followed by the MLPs and another up-module, the aggregated local region feature can be further used to reproduce the new point features with better structure details.

\begin{figure}[!t]
  \centering
  \includegraphics[width=\columnwidth]{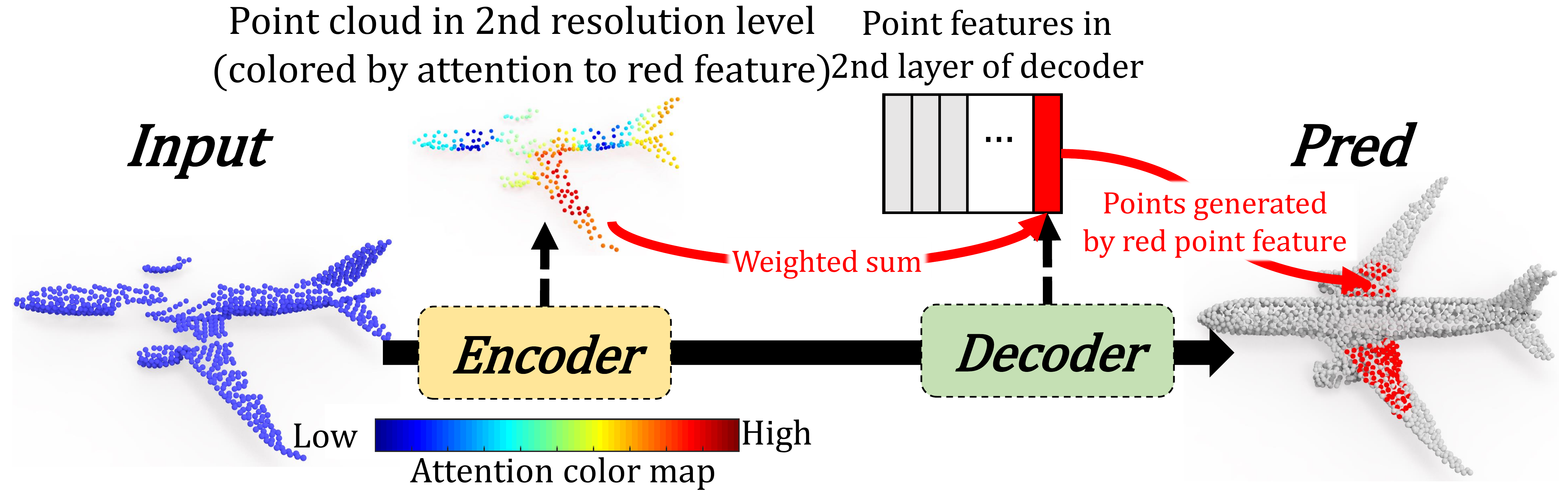}
  \caption{Illustration of the skip-attention. The skip-attention calculates the pattern similarity between local regions of complete point cloud (red points in \emph{Pred}, which is generated by red feature) and the local regions of incomplete one. The similar local regions in the incomplete point clouds is selectively fused into the decoder with attentioned weighted sum.
  }
  \label{fig:atteion_illu}
\end{figure}

\subsection{Skip-attention}
\label{sec:skip_att}
The skip-attention serves as the pipeline to communicate the local region features extracted by encoder with the point features generated by decoder. It also interprets how the network completes shapes using information from incomplete ones.
The skip-attention is designed for two purposes.
First, when generating points that are located in the existing region of incomplete inputs, the skip-attention should fuse the feature of the same region from the encoder into the decoder, and guide the decoder to reconstruct more consistent structure details in such region.
Second, when generating points that are located at the missing region of input, the skip-attention should search for referable similar regions in the original input point clouds, and guide the decoder to incorporate the shape of these similar regions as reference for inferring the shapes of missing regions.
Both of the above purposes are achieved through an attention mechanism, as shown in Figure \ref{fig:atteion_illu}, where the semantic relativeness between point features in decoder and local region features in encoder are measured by attention scores, with the higher scores indicating the more significant pattern similarity (the wings of airplane). Then, the local region features are fused into point feature by weighted sum, and finally used for predicting related regions (also the wings of plane) in the complete point cloud.

There are different possible ways to calculate attentions for the skip-attention pipeline. In this paper, we do not explore the whole space but typically choose two straightforward implementation, which work well in SA-Net.
The first one in skip-attention is to directly adopt the learnable attention mechanism as described in the up-module. And the second one is to calculate the cosine similarity as the attention measurement between features. Compare with learnable attention, the unsmoothed (no softmax activation) cosine attention brings in more information from the previous encoder network, which can establish a strong connection between point features in decoder and local region features in encoder. On the other hand, the smoothed learnable attention can preserve more information from the original point features. For learnable attention, the attention score in the $i$-th resolution level is computed between the point feature $\bs{p}_j^i$ from decoder and all of the local region features $\{\bs{r}^i_k|k=1,2,...,N^E_i\}$ from encoder, given as
\begin{equation}\small
  a^{\rm L}_{j,k} = \frac{\exp({\rm M}(\bs{p}^i_j|\theta^{\rm L}_h)^{\rm T}\cdot {\rm M}(\bs{r}^i_k|\theta^{\rm L}_l))}{\sum_{n=1}^{N_i}\exp({\rm M}(\bs{p}^i_j|\theta^{\rm L}_h)^{\rm T}\cdot {\rm M}(\bs{r}^i_n|\theta^{\rm L}_l))},
\end{equation}
where the superscript ${\rm L}$ denotes the word \emph{learnable}. For cosine distance, the attention score is given as
\begin{equation}\small
a^{\rm C}_{j,k} = \frac{(\bs{r}^i_k)^{\rm T} \bs{p}_j^i}{\|\bs{r}^i_k\|_2 \|\bs{p}_j^i\|_2},
\end{equation}
where the superscript ${\rm C}$ denotes the word \emph{cosine}. Same as the self attention in up-module, we fuse the weighted sum of local region features $\{\bs{r}^i_k\}$ into the point feature $\bs{p}_j^i$ using element-wise addition, which is the same as Eq. (\ref{eq:fuse}). In ablation study (Sec. \ref{sec:ablation}), we will quantitatively compare the performance of these two attentions.

\begin{table*}[!t]\small
\centering
\caption{Point cloud completion comparison on ShapeNet dataset in terms of per point Chamfer distance $\times 10^{4}$ (lower is better).}
\begin{tabular}{l|c|cccccccc}
\toprule
Methods &Average  &Plane    &Cabinet  &Car   &Chair   &Lamp   &Couch    &Table    &Watercraft      \\ \midrule
AtlasNet \cite{groueix2018atlasnet}   &17.69   &10.37    &23.4    &13.41    &24.16    &20.24    &20.82    &17.52    &11.62   \\
PCN  \cite{yuan2018pcn}   &14.72 &8.09    &18.32    &10.53    &19.33    &18.52    &16.44    &16.34    &10.21   \\
FoldingNet  \cite{yang2018foldingnet}   &16.48  &11.18    &20.15    &13.25    &21.48    &18.19    &19.09    &17.8    &10.69   \\
TopNet  \cite{tchapmi2019topnet}  &9.72  &5.5    &12.02    &8.9    &12.56    &\textbf{9.54}    &12.2    &\textbf{9.57}    &7.51   \\ \midrule
SA-Net(Ours)   &\textbf{7.74}  &\textbf{2.18}    &\textbf{9.11}    &\textbf{5.56}    &\textbf{8.94}    &9.98    &\textbf{7.83}    &9.94    &\textbf{7.23}   \\
\bottomrule
\end{tabular}
\label{table:shapenet_complete}
\end{table*}



\subsection{Training}
During training, the Chamfer distance (CD) $\mathcal{L}_{CD}$ and the Earth Mover distance (EMD) $\mathcal{L}_{EMD}$ are adopted as the optimization losses.
 The total loss for training is the weighted sum of the CD and EMD, defined as
 \begin{equation}\rm\small
  \mathcal{L}_{total}=\mathcal{L}_{EMD}+\lambda \mathcal{L}_{CD},
\end{equation}
where $\lambda$ is the weight parameter fixed to 10 for the experiments in our paper. The definition of $\mathcal{L}_{CD}$ and $\mathcal{L}_{EMD}$ will be detailed in \textbf{Supplementary}.

\begin{figure*}[!t]
  \centering
  \includegraphics[width=.9\textwidth]{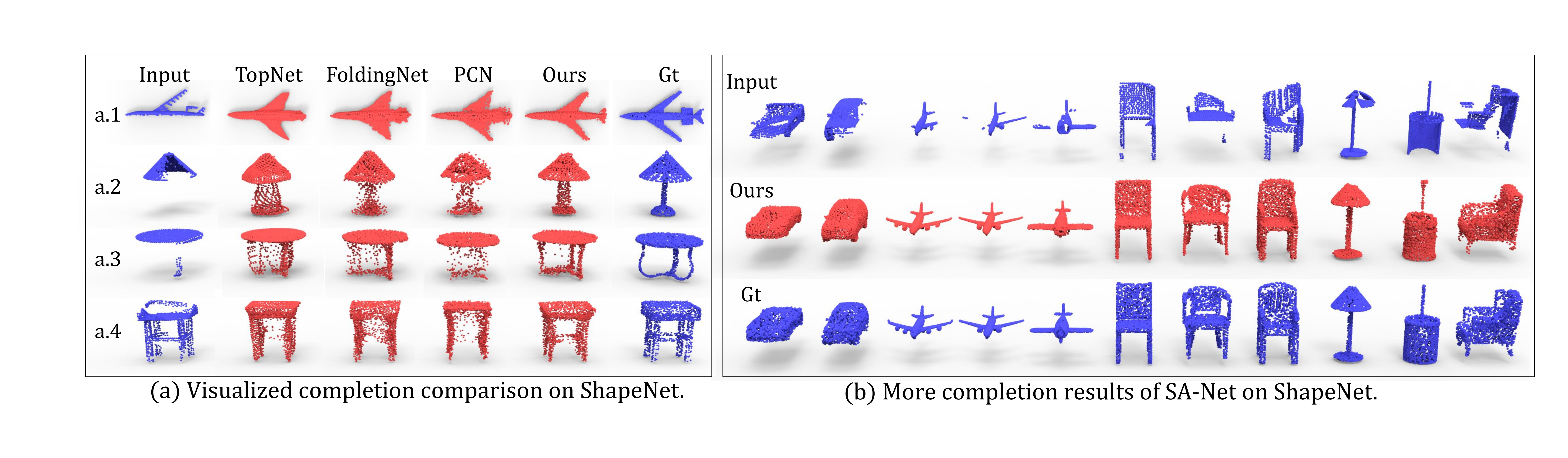}
  \caption{Visualization of point cloud completion comparison on ShapeNet dataset. We compare SA-Net with other methods in (a), and in (b) we show more completion results of SA-Net.
  }
  \label{fig:shapenet_visual}
\end{figure*}

\section{Experiments}
By default, we use the cosine similarity based skip-attention for all experiments. In Sec. \ref{sec:ablation}, we compare it with learnable attention. During evaluation, we mainly use Chamfer distance as the measurement to compare the predicted point clouds with the ground truth.
\subsection{Evaluation of Completion Performance}
\textbf{Datasets.}
To evaluate the performance of SA-Net, we conduct experiments on two large scale datasets for point cloud completion. For quantitative comparison, we follow \cite{yuan2018pcn} to evaluate our methods on ShapeNet dataset \cite{chang2015shapenet}, and generate 8 partial point clouds for each object by back-projecting 2.5D depth images from 8 views into 3D. Unlike Render4Completion \cite{hu2019render4completion}, we follow \cite{tchapmi2019topnet} to evaluate on the sparse input, which is more close to the real-world scenarios. We uniformly sample only 2,048 points on the mesh surfaces for both the complete and partial shapes.
We also qualitatively evaluate SA-Net on KITTI dataset \cite{Geiger2013IJRR}, since there is no ground truth for the incomplete shape of cars in KITTI.

\textbf{ShapeNet dataset.} We use the per point Chamfer distance as the evaluation metric. In Table \ref{table:shapenet_complete}, SA-Net is compared with two point cloud completion methods PCN \cite{yuan2018pcn} and TopNet \cite{tchapmi2019topnet}. The reconstruction based unsupervised representation learning methods FoldingNet \cite{yang2018foldingnet} and AtlasNet \cite{groueix2018atlasnet} are also included, since their basic encoder-decoder framework can also be generalized to point cloud completion task.
The results of the above 4 methods are cited from \cite{tchapmi2019topnet}.
The comparison shows that SA-Net outperforms the other methods on 6 out of 8 categories, and also achieves the best average Chamfer distance.

In Figure \ref{fig:shapenet_visual}, we show the visualization results of point cloud completion using SA-Net and compare it with the other methods, from which we can find that SA-Net predicts more reasonable shape, while preserving more consistent geometric shapes for the existing parts.
For example, in Figure \ref{fig:shapenet_visual}(a.2) and \ref{fig:shapenet_visual}(a.3), when predicting the missing lamp holders and table legs, the SA-Net generates more realistic shapes compared with the other three methods, and the points generated by SA-Net are arranged more tightly and shaped more close to the ground truth. In Figure \ref{fig:shapenet_visual}(a.1) and \ref{fig:shapenet_visual}(a.4), the SA-Net preserves the shapes of wings and the beam more consistently compared with the other three methods.
The quantitative and qualitative improvements in shape completion task prove the effectiveness of skip-attention for introducing local region features, and the ability of structure-preserving decoder for utilizing the local region features to reconstruct completion point clouds.
Moreover, in Table \ref{table:num_param}, we compare the number of trainable parameters in network of different methods, which shows that SA-Net has the least number of parameters, while achieving significantly better performance.

\begin{table}[!h]\small
\centering
\caption{The number of trainable parameters in each method.}
\resizebox{\columnwidth}{!}{\begin{tabular}{lcccc}
\toprule
Methods    &TopNet \cite{tchapmi2019topnet}  &PCN \cite{yuan2018pcn} &FoldingNet \cite{yang2018foldingnet} &SA-Net(Ours)  \\ \midrule
Params ($\times 10^6$)  &9.97   &5.29  &2.40 &\textbf{1.67}\\

\bottomrule
\end{tabular}}
\label{table:num_param}
\end{table}

\textbf{KITTI dataset.}
\begin{figure}[!t]
  \centering
  \includegraphics[width=\columnwidth]{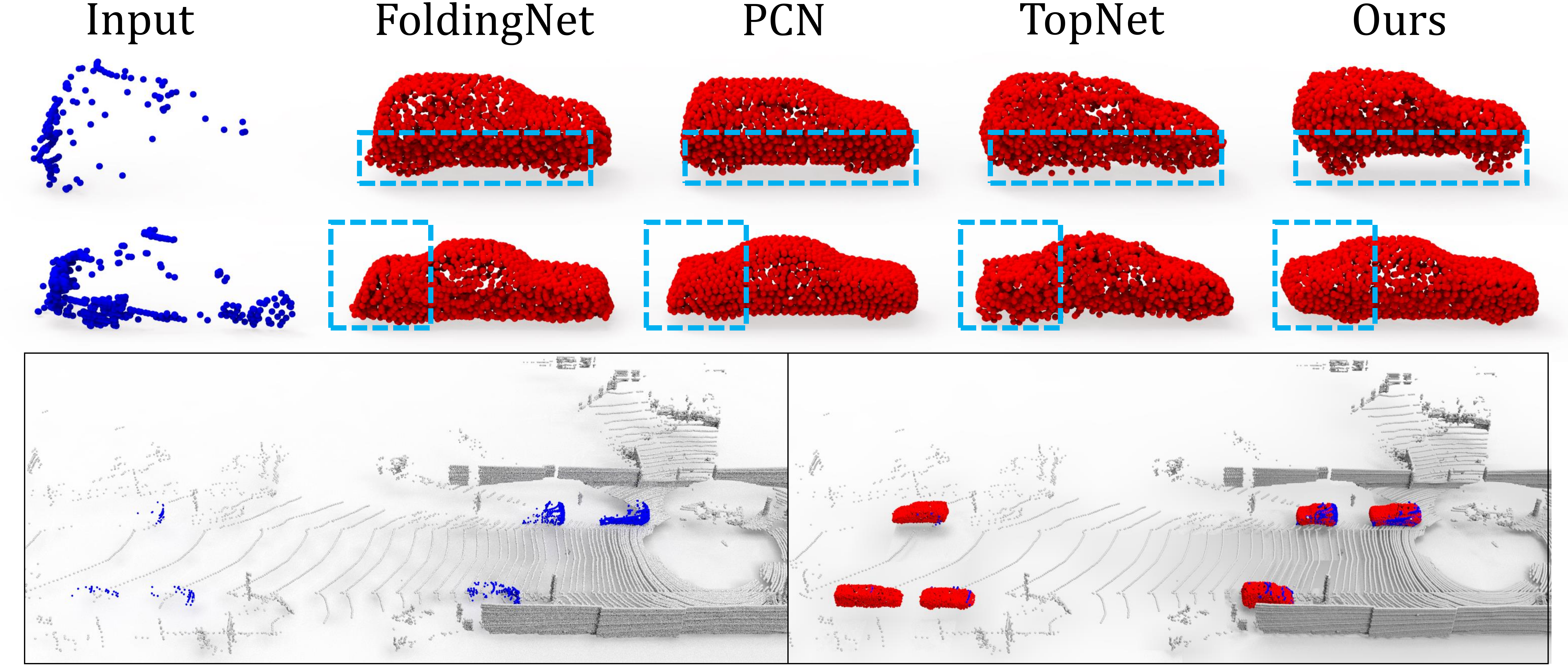}
  \caption{Visualization of the completion results on KITTI dataset.
  }
  \label{fig:kitti_visual}
\end{figure}
The KITTI dataset is collected from the real-world LiDAR scans, where the ground truth is missing for quantitative evaluation. Therefore, we qualitatively evaluate the performance of SA-Net by the visualization results. The complete cars are predicted using the parameters trained under car category in ShapeNet dataset for all methods in Figure \ref{fig:kitti_visual}. Note that in KITTI dataset, the point number of incomplete car has a large range of variation. In order to obtain a fixed point number of input, for the incomplete cars with more than 2048 points, we randomly choose 2,048 points, otherwise, we randomly select points from the input to make up to the 2,048 points. The results are shown in Figure \ref{fig:kitti_visual}, from which we can find that our SA-Net predicts more structure details (car tiers) and shapes of higher quality (car trunks).

\subsection{Ablation Study}
\label{sec:ablation}
In this subsection, we analysis the effect of important modules and hyper-parameters to SA-Net. All studies are typically conducted on the plane category for convenience.

\textbf{Effect of attention.}
We developed three variations of SA-Net to verify the effectiveness of attention in SA-Net: (1) ``No-skip'' is the variation that removes the skip-attention from the SA-Net. (2) ``Skip-L'' is the variation that replaces the cosine attention in skip-attention by the learnable attention. (3) ``Fold-C'' is the variation that replaces the learnable attention by the cosine similarity in the self-attention of folding block. All three variations have the same structure as SA-Net except for the removed/replaced module.
The results are shown in Table \ref{table:part_effects}, in which the original SA-Net achieves the best performance. The experimental results prove the effectiveness of attention used in SA-Net. The performance drop for replacing the attention in skip-attention (Skip-L) and self-attention (Fold-C) can be dedicated to the different design purposes for the two modules. The skip-attention aims to incorporate the local region features, and the unsmoothed cosine similarity allows more information to be fused into decoder. In contrast, the self-attention aims to learn a discriminative point features instead of simply merging the neighborhood features, therefore, smoothed weights (by softmax) in self-attention is more desirable for the network to preserve the original information of the point features. We specially note that, since removing the decoder of multi-resolution levels will also change the linkages of skip-attention, in Sec \ref{sec:app_analysis}, we will instead evaluate the effectiveness of decoder on the task of unsupervised shape classification.
\begin{table}[!h]\small
\centering
\caption{The effect of each module to SA-Net (plane category).}
\begin{tabular}{lcccc}
\toprule
Methods    &No-skip   &Skip-L  &Fold-C  &SA-Net \\ \midrule
CD ($\times 10^4$) &2.31   &2.25  &2.34    &\textbf{2.18}\\
\bottomrule
\end{tabular}
\label{table:part_effects}
\end{table}

\textbf{Effect of optimization loss.} To evaluate the effect of EMD loss and CD loss to SA-Net, we developed two variations: (1) ``SA-Net-EMD'' is the variation of SA-Net that is only trained using EMD loss; (2) ``SA-Net-CD'' is the variation that is only trained with CD loss. The comparison results are shown in Table \ref{table:loss_effects}, which proves that both EMD and CD contribute to the performance of SA-Net.

\begin{figure}[!t]
  \centering
  \includegraphics[width=\columnwidth]{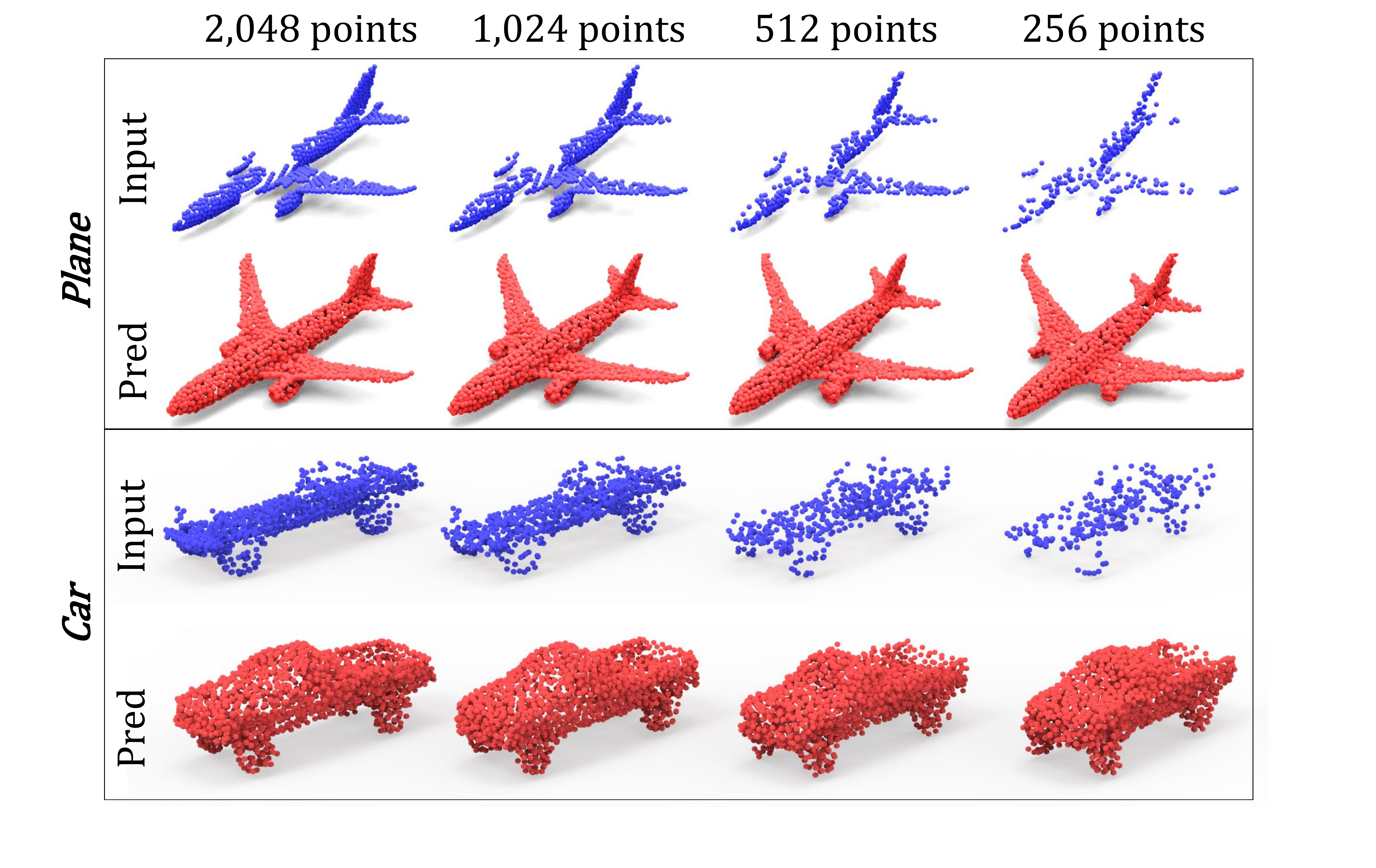}
  \caption{Visualization of completion results on different resolutions of input.
  }
  \label{fig:various_points}
\end{figure}

\begin{table}[!h]\small
\centering
\caption{The effect of each optimization loss (plane category).}
\begin{tabular}{lccc}
\toprule
Methods    &SA-Net-EMD    &SA-Net-CD   &SA-Net \\ \midrule
CD ($\times 10^4$) &2.39   &2.23    &\textbf{2.18}\\
EMD ($\times 10^2$) &3.06   &4.58  &\textbf{3.02} \\
\bottomrule
\end{tabular}
\label{table:loss_effects}
\end{table}

\textbf{Effect of input point number.}
We analyze the robustness of SA-Net on various resolutions of inputs, especially for the performance on sparse input. In this experiment, we fix the number of output point clouds to 2,048, and evaluate the performance of SA-Net on the input point clouds with resolutions ranging from 256 to 2,048. For the point size less than 2,048, we use the same strategy in KITTI dataset to randomly select points from input, and lift the number of points up to 2,048. The model performance in terms of per point CD is reported in Table \ref{table:various_points}.
In Figure \ref{fig:various_points}, we visualize the completion quality under different point number of incomplete point clouds, in which the SA-Net shows a robust performance on all input resolutions.

\begin{table}[!h]\small
\centering
\caption{The effect of input point number (plane category).}
\begin{tabular}{lcccc}
\toprule
\#Points    &2048   &1024  &512  &256  \\ \midrule
CD ($\times 10^4$)  &\textbf{2.18}   &2.28  &2.45 &3.31\\

\bottomrule
\end{tabular}
\label{table:various_points}
\end{table}

\begin{figure}[!t]
  \centering
  \includegraphics[width=\columnwidth]{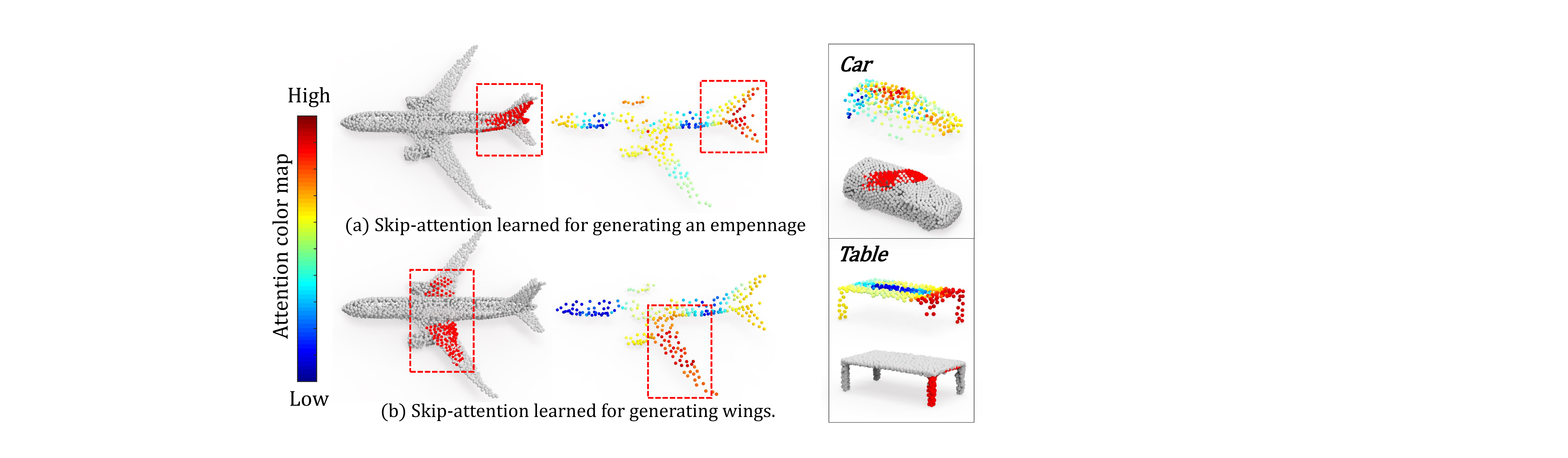}
  \caption{Visualization of the attention learned in skip-attention.
  }
  \label{fig:att_visual}
\end{figure}

\textbf{Visualization of skip-attention.}
In Figure \ref{fig:att_visual}, we visualize the attention in the second resolution level of decoder, which is to predict a complete plane. We compare the skip-attention learned for generating the empennage and part of the two wings. The points generated by the same point feature are colored by red in the left half of Figure \ref{fig:att_visual}(a) and \ref{fig:att_visual}(b), and the corresponding attention scores that point feature assigned to the local regions of incomplete point cloud are visualized in the right half. As shown in Figure \ref{fig:att_visual}(a), when generating points that belong to the empennage, the skip-attention searches for relative local regions (which is also the empennage) in the input point clouds for prediction. In Figure \ref{fig:att_visual}(b), when predicting the points of wings (where the right wing are missing), the skip-attention selects the region of left wing (by assigning higher attention) in incomplete point cloud for predicting the shape of both wings.  Similar pattern is also observed on other categories as shown in Figure \ref{fig:att_visual}.

\begin{figure}[!t]
  \centering
  \includegraphics[width=\columnwidth]{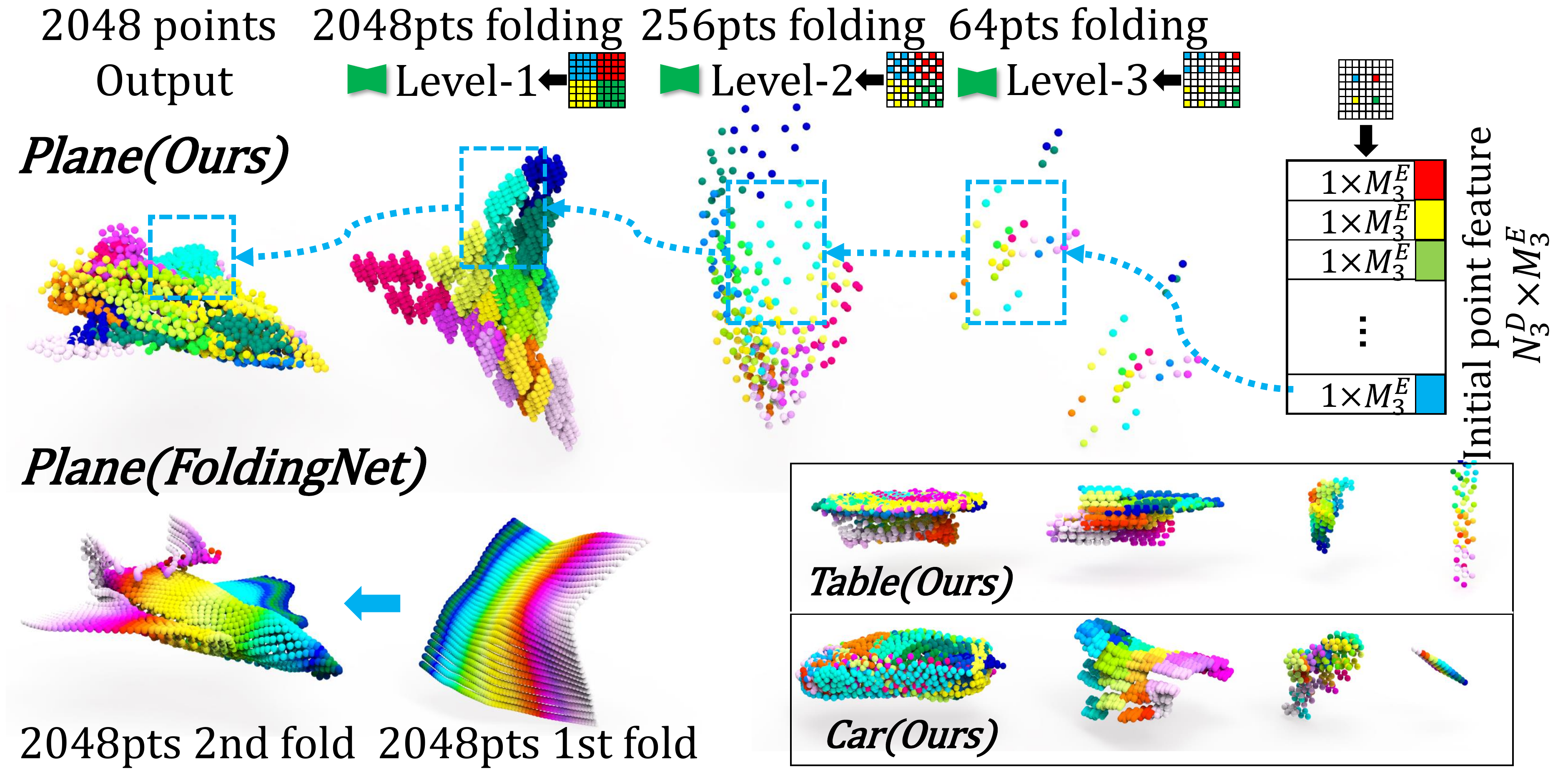}
  \caption{Visualization of the hierarchical folding in each level of decoder. We track the folding and point number expansion process of a specific initial vector colored by blue, and illustratively show the 2D grids sampling process.
  }
  \label{fig:folding_visual}
\end{figure}

\textbf{Visualization of hierarchical folding.}
In Figure \ref{fig:folding_visual}, we visualize the hierarchical folding in decoder. We track the folding process of a specific vector colored by blue, and denote the points derived from this blue vector with blue rectangular in each level. From a local perspective, we observe that each initial point feature successfully learns to generate a specific region on the plane. And in the case of blue initial point feature, it generates the left wing of the plane. On the other hand, from a global perspective, we can observe that the folding process of SA-Net does not restrictively follow the 2D manifold assumption like FoldingNet \cite{yang2018foldingnet}. As pointed out by \cite{tchapmi2019topnet}, enforcing learning from the 2D manifold structure may not be optimal for training, because the space of possible solutions is constrained. Therefore, the subtle deviation from 2D manifold, which is observed in SA-Net, is more flexible for learning to generate variant shapes and preserve better structure details. Both of the observations prove the effectiveness of hierarchical folding. In addition, we also visualize the folding process under car and table categories in Figure \ref{fig:folding_visual}.

\begin{figure*}[!t]
  \centering
  \includegraphics[width=\textwidth]{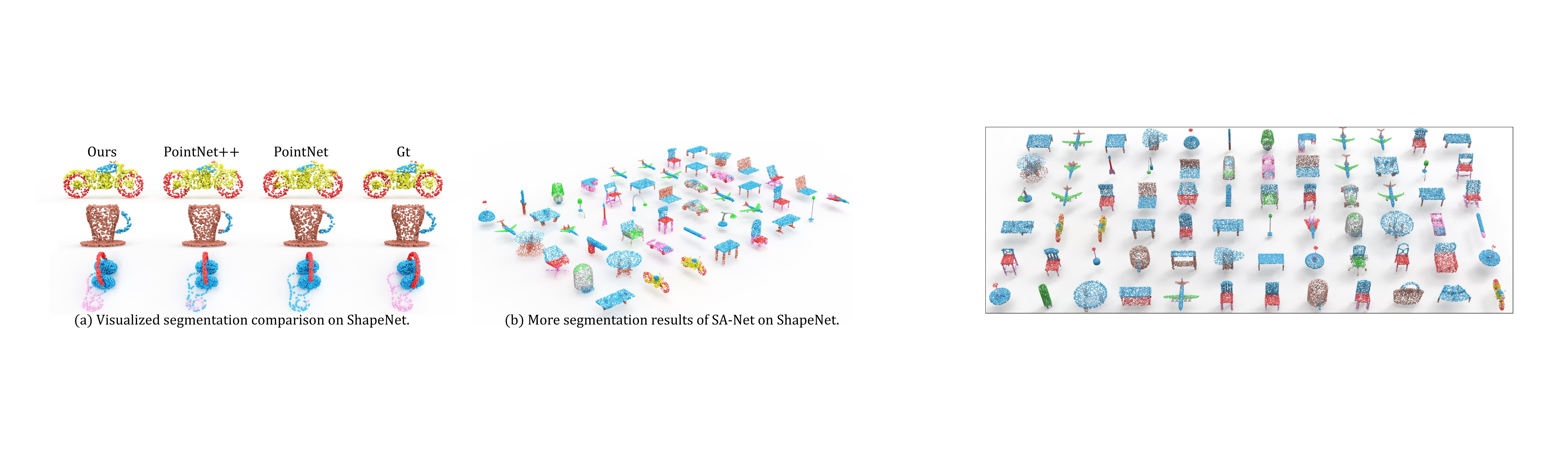}
  \caption{Segmentation visualization on ShapeNet. We compare SA-Net with baseline PointNet and PointNet++ in (a). In (b), we show more segmentation results of SA-Net. Note that there is no correspondence between colors and labels across object categories in (b).
  }
  \label{fig:shapenet_seg}
\end{figure*}

\subsection{Model Analysis on Applications}
\label{sec:app_analysis}
\textbf{Skip-attention for semantic segmentation.}
To further verify the effectiveness of skip-attention proposed in Sec \ref{sec:skip_att}, we conduct the semantic segmentation experiment on the ShapeNet dataset \cite{yi2016a}, where the dataset splittings follow the previous method of PointNet++ \cite{qi2017pointnet++}. The segmentation variation (SA-Net-seg) of SA-Net uses exactly the same architecture as PointNet++, except for the skip-attention connecting the local region features in encoder with the features in interpolation layers. The comparison in terms of part-averaged intersection over union (pIoU, $\%$) and mean per-class pIoU (mpIoU, $\%$) \cite{li2018pointcnn} is shown in Table \ref{table:shapenet_seg}, from which we can find that SA-Net-seg drastically improves the segmentation performance compared with the baseline method of PointNet++. Specifically, the skip-attention improves the performance of backbone PointNet++ by $0.6\%$ in terms of mIoUs.
In Figure \ref{fig:shapenet_seg}(a), we visualize the segmentation results and compared SA-Net-seg with the baseline PointNet and PointNet++, from which we can find that SA-Net-seg yields more precise prediction of semantic labels. Especially, the SA-Net-seg significantly improves the segmentation accuracy on the tier of the motorcycle, where the body and the tier are heavily overlapped with each other.
Such improvement results from the local region features conveyed by skip-attention from the encoder, which helps the interpolation layers make more discriminative prediction in the local regions. Figure \ref{fig:shapenet_seg}(b) gives more segmentation results.
\begin{table}[!h]\small
\centering
\caption{Semantic segmentation results ($\%$) on ShapeNet.}
\begin{tabular}{lcc}
\toprule
Methods    &pIoU   &mpIoU      \\ \midrule
PointNet \cite{qi2017pointnet} &83.7   &80.4  \\
PointNet++ \cite{qi2017pointnet++} &85.1 &81.9   \\
SO-Net \cite{li2018so}   &84.9   & 81.0\\
DGCNN\cite{wu2018dgcnn} &85.1   &82.3 \\
PointCNN \cite{li2018pointcnn}   &\textbf{86.1}   & \textbf{84.6}\\
\midrule
SA-Net-seg(Ours)    &85.7  &83.0  \\
\bottomrule
\end{tabular}
\label{table:shapenet_seg}
\end{table}

\begin{table}[tp]\small
\centering
\caption{The classification comparison under ModelNet40.}
\label{table:modelent_cls}
\begin{tabular}{lccc}
\toprule
Methods          &Supervised &Accuracy(\%)  \\ \midrule
 PointNet\cite{qi2017pointnet}        &Yes  &89.2        \\
 PointNet++ \cite{qi2017pointnet++}    &Yes  &90.7         \\
 PointCNN\cite{li2018pointcnn}     &Yes  &92.2         \\
 DGCNN\cite{wu2018dgcnn}        &Yes  &92.2         \\
 SO-Net\cite{li2018so}         &Yes  &90.9      \\\midrule
 LGAN\cite{achlioptas2018learning}        &No   &85.7    \\
 LGAN\cite{achlioptas2018learning}(MN40)      &No   &87.3    \\
 FoldingNet\cite{yang2018foldingnet}          &No   &88.4    \\
 FoldingNet\cite{yang2018foldingnet}(MN40)      &No   &84.4    \\
 MAP-VAE\cite{han2019multi}      &No     &90.2         \\
 L2G\cite{liu2019l2g}       &No   &\textbf{90.6} \\\midrule
 SA-Net-cls(Ours)             &No   &\textbf{90.6} \\
\bottomrule
\end{tabular}
\end{table}

\textbf{Structure-preserving decoder for unsupervised representation learning in shape classification.}
In order to verify the effectiveness of our structure-preserving decoder, we further conduct unsupervised shape classification experiments on ModelNet40 \cite{wu20153d}. The training and test settings on ModelNet40 also follow the PointNet++ \cite{qi2017pointnet++}. In this experiment, we use a classification variation (SA-Net-cls) of SA-Net, in which we remove the skip-attention from SA-Net.
The reason is that we use the global representation for predicting class label by a support vector machine (SVM), and remove the skip-attention can enhance the information embedded in the global representation, since it forces the decoder to decode a whole point cloud only based on the single global representation.
The encoder and decoder are trained by reconstructing itself.
In Table \ref{table:modelent_cls}, we compare the classification performance of SA-Net-cls with counterpart methods, where all results are obtained under 1,024 points input without normal vector. From Table \ref{table:modelent_cls} we can find that our SA-Net-cls achieves the best performance in the unsupervised learning methods. The result of SA-Net-cls is also comparable with the supervised methods. Especially, we note that the classification accuracy of our SA-Net-cls is only $0.1\%$ lower than the supervised PointNet++, which is exactly the same backbone used as our encoder.

\section{Conclusion}
We propose a novel Skip-Attention Network (SA-Net) for point cloud completion. Through the proposed skip-attention, SA-Net can effectively utilize the features of local regions in input point clouds for completion task. In order to exploit the local regions at different resolutions, the structure-preserving decoder is further proposed to progressively generate point clouds, and incorporate local region features at different resolutions. The completion experiments on ShapeNet and KITTI prove the effectiveness of SA-Net. The segmentation and classification experiments on ShapeNet and ModelNet40 further demonstrate the effectiveness of skip-attention and structure-preserving decoder, respectively.

\bibliographystyle{ieee}
\bibliography{sanet}

\end{document}